\documentclass[11pt,a4paper]{article}
\usepackage[hyperref]{emnlp2020}
\usepackage{times}
\usepackage{latexsym}

\usepackage{amsthm,amsmath,amssymb}
\usepackage{mathrsfs}
\usepackage{graphicx}
\usepackage{multirow}
\usepackage{multicol}
\usepackage{longtable}
\usepackage{makecell}
\usepackage{array}

\usepackage{microtype}

\aclfinalcopy 
\def\aclpaperid{1500} 

\title{Unsupervised Extractive Summarization by Pre-training Hierarchical Transformers}

\author{Shusheng Xu$^{1}$\thanks{\hspace{0.1cm}~Work done during the first author's internship at \mbox{Microsoft} Research Asia.}~, Xingxing Zhang$^{2}$, Yi Wu$^{1,3}$, Furu Wei$^2$ \and Ming Zhou$^{2}$\\[0.5ex]
	$^1$ IIIS, Tsinghua Unveristy, Beijing, China \\
	$^2$ Microsoft Research Asia, Beijing, China \\
	$^3$ Shanghai Qi Zhi institute, Shanghai China\\
	[0.5ex]
	{\tt xuss20@mails.tsinghua.edu.cn} \\
	{ \tt\{xizhang,fuwei,mingzhou\}@microsoft.com} \\ 
	{\tt jxwuyi@gmail.com}
}

\date{}

\begin{document}
\maketitle
\begin{abstract}

Unsupervised extractive document summarization aims to select important sentences from a document without using labeled summaries during training. Existing methods are mostly graph-based with sentences as nodes and edge weights measured by sentence similarities. In this work, we find that transformer attentions can be used to rank sentences for unsupervised extractive summarization. Specifically, we first pre-train a hierarchical transformer model using unlabeled documents only. Then we propose a method to rank sentences using sentence-level self-attentions and pre-training objectives. Experiments on CNN/DailyMail and New York Times datasets show our model achieves state-of-the-art performance on unsupervised summarization.  We also find in experiments that our model is less dependent on sentence positions. When using a linear combination of our model and a recent unsupervised model explicitly modeling sentence positions, we obtain even better results.
\end{abstract}

\section{Introduction}

Document summarization is the task of transforming a long document into its shorter version while still retaining its important content. Researchers have explored many paradigms for summarization, while the most popular ones are \emph{extractive} summarization and \emph{abstractive} summarization \cite{Nenkova:McKeown:2011}. As their names suggest, extractive summarization generates summiries by extracting text from original documents, and abstractive summarization rewrites documents by paraphrasing or deleting some words or phrases.

Most summarization models require labeled data, where documents are paired with human written summaries. Unfortunately, human labeling for summarization task is expensive and therefore high quality large scale labeled summarization datasets are rear \cite{hermann:2015:nips} compared to growing web documents created everyday. It is also not possible to create summaries for documents in all text domains and styles. In this paper, we focus on unsupervised summarization, where we only need unlabeled documents during training.

Many attempts for unsupervised summarization are extractive \cite{carbonell1998use,radev-etal-2000-centroid,lin2002single,mihalcea2004textrank,erkan2004lexrank,wan-2008-exploration,wan2008multi,hirao-etal-2013-single,parveen-etal-2015-topical}. The core problem is to identify salient sentences in a document. 
The most popular approaches among these work rank sentences in the document using graph based algorithms, where each node is a sentence and weights of edges are measured by sentence similarities. Then a graph ranking method is employed to estimate sentence importance. For example, TextRank \cite{mihalcea2004textrank} utilizes word co-occurrence statistics to compute similarity and then employs PageRank \cite{page1997pagerank} to rank sentences. Sentence similarities in \cite{zheng-lapata-2019-sentence} are measured with BERT \cite{devlin:2019:naacl} and sentences are sorted w.r.t. their centralities in a directed graph.

Recently, there has been increasing interest in developing unsupervised abstractive summarization models \cite{wang-lee-2018-learning,fevry-phang-2018-unsupervised,Chu2019MeanSumAN,Yang2020TEDAP}. These models are mostly based on sequence to sequence learning \cite{sutskever2014sequence} and sequential denoising auto-encoding \cite{dai2015semi}. Unfortunately, there is no guarantee that summaries produced by these models are grammatical and consistent with facts described original documents.

%

%

\newcite{zhang:2019:acl} propose an unsupervised method to pre-train a hierarchical transformer model (i.e., HIBERT) for document modeling. The hierarchical transformer has a token-level transformer to learn sentence representations and a sentence-level transformer to learn interactions between sentences with self-attention.  In \newcite{zhang:2019:acl}, \mbox{HIBERT} is applied to supervised extractive summarization. However, we believe that after pre-training HIBERT on large scale unlabeled data, the self-attention scores in the sentence-level transformer becomes meaningful for estimating the importance of sentences. Intuitively, if many sentences in a document attend to one particular sentence with high attention scores, then this sentence should be important. 
In this paper, we find that (sentence-level) transformer attentions (in a hierarchical transformer) can be used to rank sentences for unsupervised extractive summarization, while previous work mostly leverage graph based (or rule based) methods and sentence similarities computed with off-the-shelf sentence embeddings. Specifically, we first introduce two pre-training tasks for hierarchical transformers (i.e., extended \mbox{HIBERT}) to obtain sentence-level self-attentions using unlabled documents only. Then, we design a method to rank sentences by using sentence-level self-attentions and pre-training objectives. Experiments on CNN/DailyMail and New York Times datasets show our model achieves state-of-the-art performance on unsupervised summarization. We also find in experiments that our model is less dependent on sentence positions. When using a linear combination of our model and a recent unsupervised model explicitly modeling sentence positions, we obtain even better results. Our code and models are available at \url{https://github.com/xssstory/STAS}.



\section{Related Work}

In this section, we introduce work on supervised summarization, unsupervised summarization and pre-training.

\paragraph{Supervised Summarization}
Most summarization models require supervision from labeled datasets, where documents are paired with human written summaries.  As mentioned earlier, extractive summarization aims to extract important sentences from documents and it is usually viewed as a (sentence) ranking problem by using  scores from classifiers \cite{kupiec1995trainable} or sequential labeling models \cite{conroy2001text}. Summarization performance of this class of  methods are greatly improved, when human engineered features \cite{radev2004mead,nenkova2006compositional,filatova-hatzivassiloglou-2004-event} are replaced with convolutional neural networks (CNN) and long short-term memory networks (LSTM) \cite{cheng-lapata-2016-neural,nallapati2017summarunner,narayan-etal-2018-ranking,zhang-etal-2018-neural-latent}.

Abstractive summarization on the other hand can generate new words or phrases and are mostly based on sequence to sequence (seq2seq) learning \cite{bahdanau:2015:iclr}. To better fit in the summarization task, the original seq2seq model is extended with copy mechanism \cite{gu-etal-2016-incorporating}, coverage model \cite{see-etal-2017-get},  reinforcement learning \cite{paulus2018a} as well as bottom-up attention \cite{gehrmann-etal-2018-bottom}. Recently, pre-trained transformers \cite{vaswani:2017:nips} achieve tremendous success in many NLP tasks \cite{devlin:2019:naacl,liu:2019:roberta}. Pre-trained methods customized for both extractive \cite{zhang:2019:acl,liu-lapata-2019-text} and abstractive \cite{dong2019unified,lewis2019bart} summarization again advance the state-of-the-art in supervised summarization. Our model also leverages pre-trained methods and models, but it is unsupervised.

\paragraph{Unsupervised Summarization}
Compared to supervised models, unsupervised models only need unlabeled documents during training. Most unsupervised extractive models are graph based \cite{carbonell1998use,radev-etal-2000-centroid,lin2002single,mihalcea2004textrank,erkan2004lexrank,wan-2008-exploration,wan2008multi,hirao-etal-2013-single,parveen-etal-2015-topical}. For example, TextRank \cite{mihalcea2004textrank} treats sentences in a document as nodes in an undirected graph, and edge weights are measured with co-occurrence based similarities between sentences. Then PageRank \cite{page:1999:pagerank} is employed to determine the final ranking scores for sentences. \newcite{zheng-lapata-2019-sentence} builds directed graph by utilizing BERT \cite{devlin:2019:naacl} to compute sentence similarities. The importance score of a  sentence is the weighted sum of all its out edges, where weights for edges between the current sentence and preceding sentences are negative. Thus, leading sentences tend to obtain high scores. Unlike \newcite{zheng-lapata-2019-sentence}, sentence positions are not explicitly modeled in our model and therefore our model is less dependent on sentence positions (as shown in experiments).

There are also an interesting line of work on unsupervised abstractive summarization. \newcite{Yang2020TEDAP} pre-trains a seq2seq Transformer by predicting the first three sentences of news documents and then further tunes the model with semantic classification and denoising auto-encoding objectives. 
The model described in \newcite{wang-lee-2018-learning} utilizes seq2seq auto-encoding coupled with adversarial training and reinforcement learning. \newcite{fevry-phang-2018-unsupervised} and \newcite{baziotis-etal-2019-seq} focus on sentence summarization (i.e., compression). \newcite{Chu2019MeanSumAN} proposes yet another denoising auto-encoding based model in multi-document summarization domain. However, the performance of these unsupervised models are still unsatisfactory compared to their extractive counterparts. 

\paragraph{Pre-training}
Pre-training methods in NLP learn to encode text by leveraging unlabeled text. Early work mostly concentrate on pre-training word embeddings \cite{Mikolov2013EfficientEO, pennington2014glove, bojanowski2017enriching}. 
Later, sentence encoder can also be pre-trained with language model (or masked language model) objectives \cite{peters-etal-2018-deep, radford2018improving,devlin:2019:naacl,liu:2019:roberta}. \newcite{zhang:2019:acl}
propose a method to pre-train a hierarchical transformer encoder (document encoder) by predicting masked sentences in a document for \emph{supervised summarization}, while we focus on \emph{unsupervised summarization}. In our method, we also propose a new task (sentence shuffling) for pre-training hierarchical transformer encoders. \newcite{iter2020pretraining} propose a contrastive pre-training objective to predict relative distances of surrounding sentences to the anchor sentence, while our sentence shuffling task predicts original positions of sentences from a shuffled docuemt. Besides,  pre-training methods mentioned above focus on learning good word, sentence or document representations for downstream tasks, while our method focuses on learning sentence level attention distributions (i.e., sentence associations), which is shown in our experiments to be very helpful for unsupervised summarization.

\section{Model}

In this section, we describe our unsupervised summarization model \textsc{Stas} (as shorthand for \textbf{S}entence-level \textbf{T}ransformer based 
\textbf{A}ttentive \textbf{S}ummarization).
We first introduce how documents are encoded in our model. Then we present methods to pre-trained our document encoder. Finally we apply the pre-trained encoder to unsupervised summarization.

\subsection{Document Modeling}
\label{sec:docmodel}

\begin{figure}[t]
\centering
\includegraphics[width=1\linewidth]{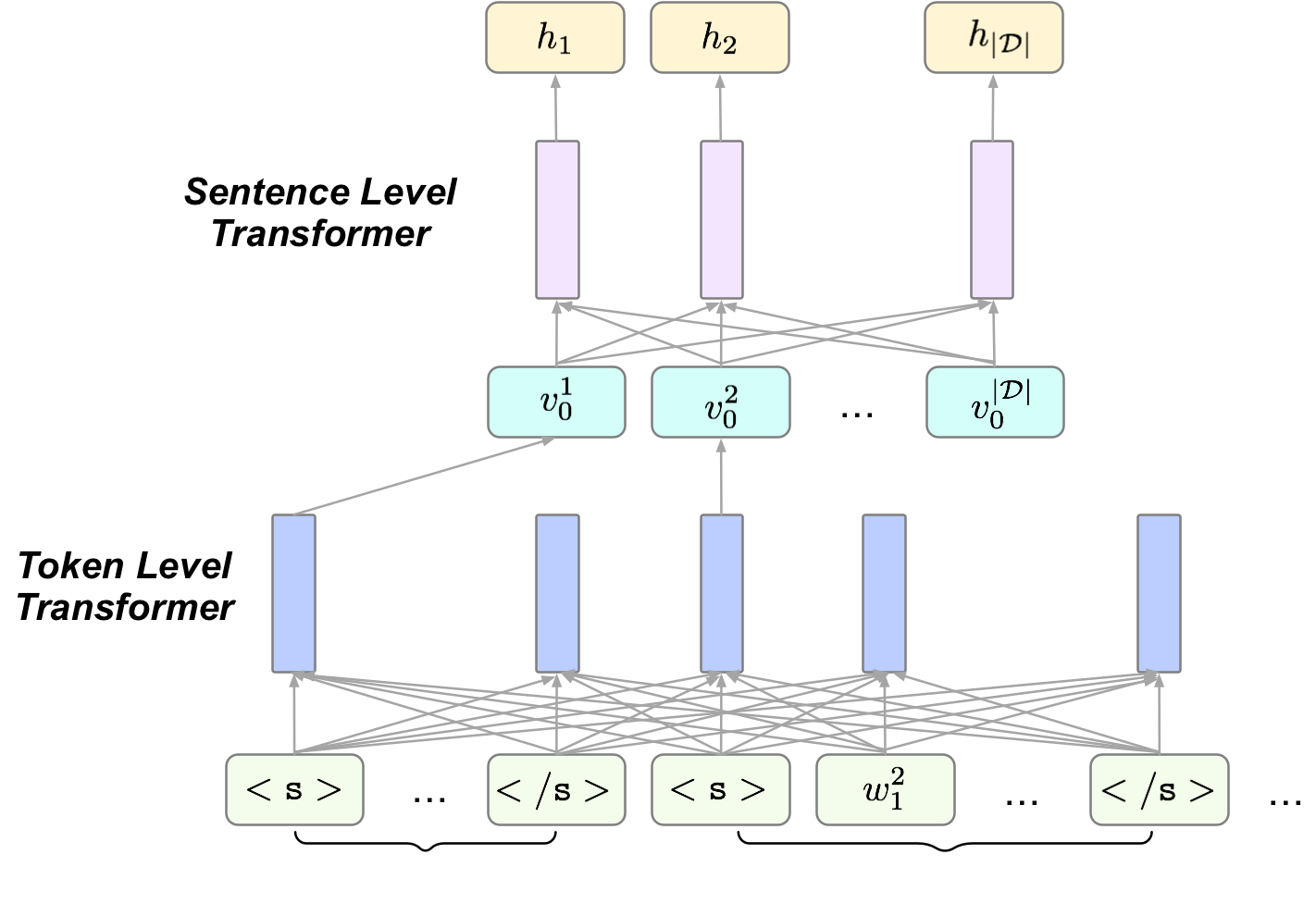}
\caption{The architecture of our hierarchical encoder, the token level Transformer encodes tokens and then the sentence level Transformer learns final sentence representations from representations at $\text{\tt <s>}$.
}  
\label{fig:encoder_architecture} 
\end{figure}

Let $\mathcal{D} = ( S_1, S_2, \dots, S_{| \mathcal{D} |} )$ denote a document, where $S_i = ( w_0^i, w_1^i, w_2^i, \dots, w_{| S_i |}^i)$ is a sentence in $\mathcal{D}$ and $w_j^i$ is a token in $S_i$. As a common wisdom, we also add two special tokens (i.e., $w_0^i=\text{\tt <s>}$ and $w_{| S_i |}^i=\text{\tt </s>}$) to $S_i$, which represents the begin and end of a sentence, respectively. Transformer models \cite{vaswani:2017:nips}, which are composed of multiple self-attentive layers and skip connections \cite{he:2016:cvpr}, have shown tremendous success in text encoding \cite{devlin:2019:naacl}. 
Due to the hierarchical nature of documents, we encode the document $\mathcal{D}$ using a hierarchical Transformer encoder, which contains a \emph{token-level} Transformer $Trans^T$ and a \emph{sentence-level} Transformer $Trans^S$ as shown in Figure \ref{fig:encoder_architecture}. Let $||$ denote an operator for sequences concatenation. $Trans^T$ views $\mathcal{D}$ as a \emph{flat} sequence of tokens denoted as $D = ( S_1 \, || \, S_2 \, || \dots || \, S_{| \mathcal{D} |} ) $. After we apply $Trans^T$ to $D$, we obtain contextual representations for all tokens $(\mathbf{v}_0^1, \mathbf{v}_1^1, \dots, \mathbf{v}_{|S_1|}^1, \dots, \mathbf{v}_j^i, \dots, \mathbf{v}^{ |\mathcal{D}| }_0, \dots, \mathbf{v}^{ |\mathcal{D}| }_{|S_{|\mathcal{D}|}|})$. We use the representation at each {\tt <s>} token as the representation for that sentence and therefore representations for all sentences in $\mathcal{D}$ are $\mathbf{V} = (\mathbf{v}_0^1, \mathbf{v}_0^2, \dots, \mathbf{v}_0^{|\mathcal{D}|})$. The \emph{sentence-level} Transformer $Trans^S$  takes $\mathbf{V}$ as input and learns sentence representations given other sentences in $\mathcal{D}$ as context:
\begin{equation}
\mathbf{H}, \mathbf{A} = Trans^S( \mathbf{V} )
\end{equation}
where $\mathbf{H}=(\mathbf{h}_1, \mathbf{h}_2, \dots, \mathbf{h}_{|\mathcal{D}|})$ and $\mathbf{h}_i$ is the final representation of $S_i$; $\mathbf{A}$ is the self-attention matrix and $\mathbf{A}_{i,j}$ is the attention score from sentence $S_i$ to sentence $S_j$. $Trans^S$ contains multiple layers and each layer contains multiple attention heads. To obtain $\mathbf{A}$, we first average the attention scores across different heads and then across different layers. 
Our hierarchical document encoder is similar to the hierarchical Transformer model described in \newcite{zhang:2019:acl}. The main difference is that our \emph{token-level} Transformer encodes all sentences in a document as a whole rather than separately.

%

\subsection{Pre-training}
\label{sec:pretraining}
In this section, we pre-train the hierarchical document encoder introduced in Section \ref{sec:docmodel} using unlabeled documents only. We expect that after pre-training, the encoder would obtain the ability of modeling interactions (i.e., attentions) among sentences in a document. In this following, we introduce two tasks we used to pre-train the encoder.

\paragraph{Masked Sentences Prediction} The first task is Masked Sentences Prediction (MSP) described in \newcite{zhang:2019:acl}. We randomly mask 15\% of sentences in a document and then predict the original sentences. Let $\mathcal{D} = (S_1, S_2, \dots, S_{|\mathcal{D}|})$ denote a document and $\widetilde{\mathcal{D}} = (\widetilde{S}_1,  \dots, \widetilde{S}_{|\mathcal{D}|})$ the document with some sentences masked, where
\begin{equation}
\widetilde{S}_i = \left\{
\begin{aligned}
& S_i & 85\% \quad \text{of cases} \\
& mask(S_i) &15\% \quad \text{of cases} \\
\end{aligned}
\right.
\end{equation}
$mask(S_i)$ is a function to mask $S_i$, which in 80\% of cases replaces each word in $S_i$ with the {\tt [MASK]} token, in 10\% of cases replaces $S_i$ with a random sentence and in the remaining 10\% of cases keep $S_i$ unchanged. The masking strategy is similar to that of BERT \cite{devlin:2019:naacl}, but it is applied on sentence level. Let $\mathcal{I} = \{ i | \widetilde{S}_i  = mask(S_i) \}$ denote the set of indices for masked sentences and $\mathcal{O}=\{ S_i | i \in \mathcal{I} \}$ the original sentences corresponding to masked sentences. 

\begin{figure}[t]
	\centering
	\includegraphics[width=1\linewidth]{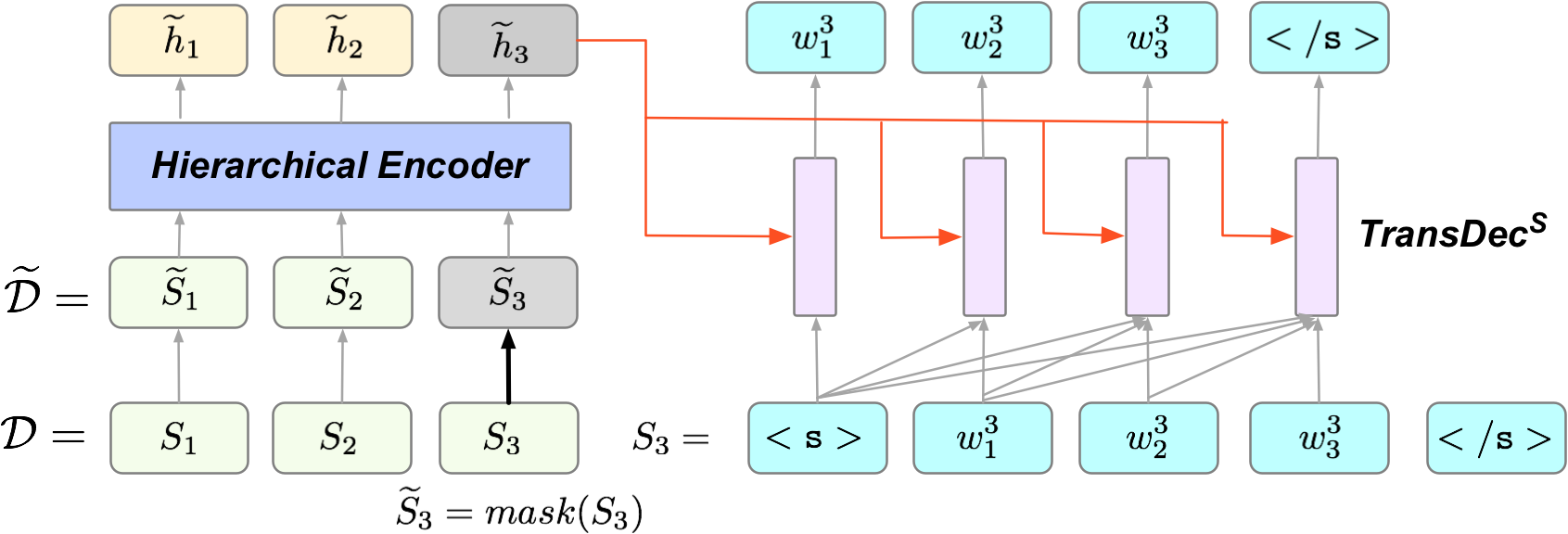}
	\caption{An example of masked sentences prediction. The third sentence in the document is masked and the hierarchical encoder encodes the masked document. We then use $TransDec^S$ to predict the original sentence one token at a time.}  
	\label{fig:MSP} 
\end{figure}

Supposing $i \in \mathcal{I}$ , we demonstrate how we predict the original sentence $S_i = (w_0^i,w_1^i,\dots,w_{|S_i|}^i)$ given $\widetilde{\mathcal{D}}$. As shown in Figure \ref{fig:MSP}, we first encode $\widetilde{\mathcal{D}}$ using the encoder in Section \ref{sec:docmodel} and obtain $\widetilde{ \mathbf{H} }=(\widetilde{ \mathbf{h} }_1, \widetilde{ \mathbf{h} }_2, \dots, \widetilde{ \mathbf{h} }_{|\mathcal{D}|})$. Then we use $\widetilde{ \mathbf{h} }_i$ (i.e., the contextual representation of $\widetilde{S}_i$) to predict $S_i$ one token at a time with a conditional Transformer decoder $TransDec^M$. We inject the information of $\widetilde{S}_i$ to $TransDec^M$ by adding $\widetilde{ \mathbf{h} }_i$ after the self attention sub-layer of each Transformer block in $TransDec^M$. Assuming $w_{0:j-1}^i$ has been generated, the probability of $w_j^i$ given $w_{0:j-1}^i$ and $\widetilde{\mathcal{D}}$ is
\begin{align}
\label{eq:prob}
\hat{ \mathbf{h} }_j^i =  TransDec^M(w_{0:j-1}^i, \widetilde{ \mathbf{h} }_i) \\
p(w_j^i | w_{0:j-1}^i, \widetilde{\mathcal{D}}) = \text{softmax}( \mathbf{W}_o \hat{ \mathbf{h} }_j^i  )
\end{align}
The probability of all original sentences given $\widetilde{\mathcal{D}}$ is
\begin{equation}
\label{eq:pro_masked}
p( \mathcal{O} |  \widetilde{\mathcal{D}} ) = \prod_{S_i \in  \mathcal{O} } \prod_{j=1}^{|S_i|} p(w_j^i | w_{0:j-1}^i, \widetilde{\mathcal{D}})
\end{equation}
MSP is proposed in HIBERT \cite{zhang:2019:acl} for \emph{supervised summarization}, while we use MSP and transformer attention for sentence ranking in \emph{unsupervised summarization} (Section \ref{sec:extract_sum}). Note that the goal and the way of using MSP in this work is different from these in HIBERT.
\paragraph{Sentence Shuffling}
We propose a new task that shuffles the sentences in a document and then select sentences in the original order one by one. We expect that the hierarchical document encoder can learn to select sentences based on their contents rather than positions.
\begin{figure}[t]
	\centering
	\includegraphics[width=1\linewidth]{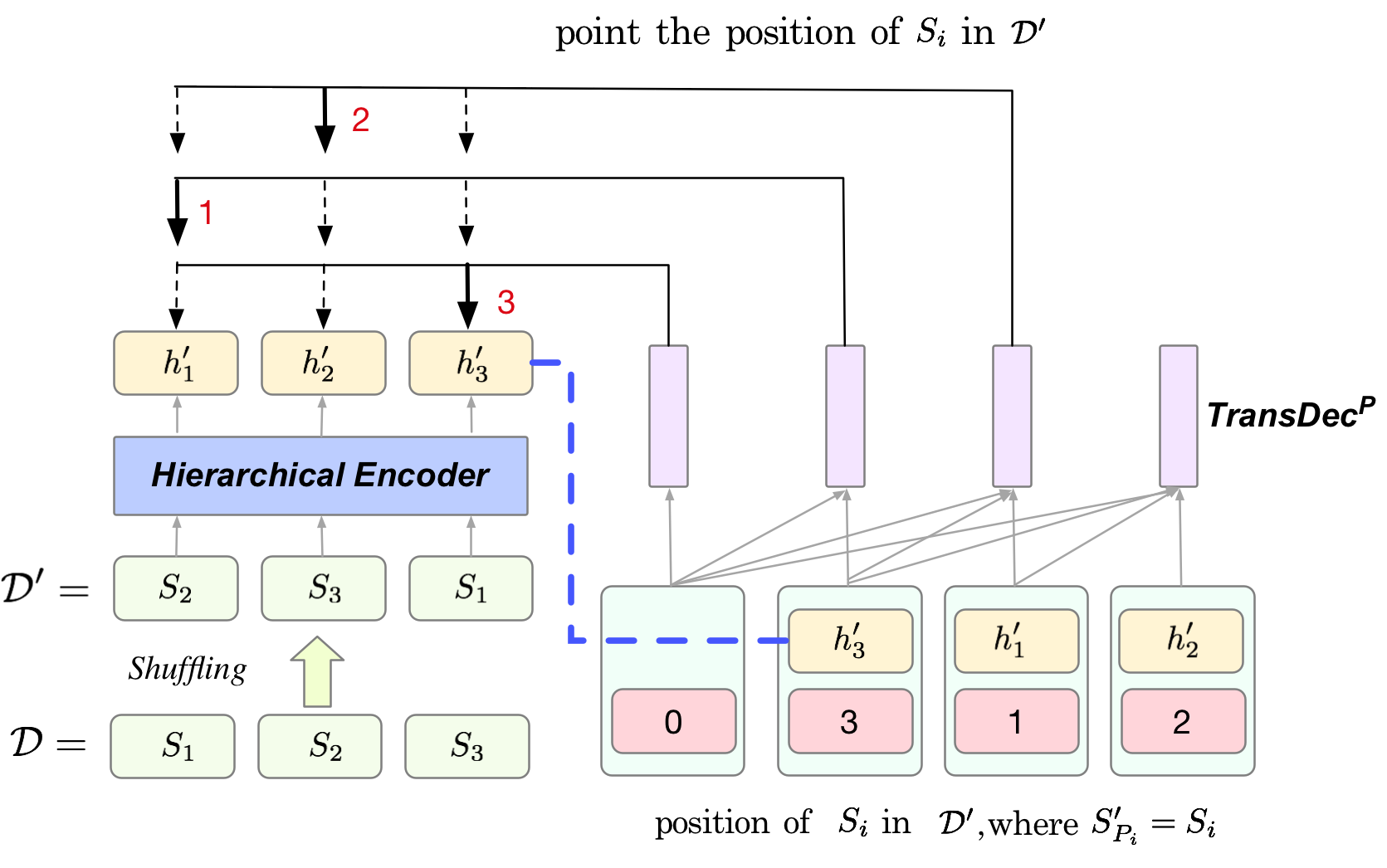}
	\caption{An example of Sentence Shuffling. The sentences in the document are shuffled and then pass through the hierarchical encoder, then a Pointer Network with $TransDec^P$ as its decoder is adopted to predict the positions of original sentences in the shuffled document.}  
	\label{fig:reordering_architecture} 
\end{figure}

Recall that $\mathcal{D} = (S_1, S_2, \dots, S_{|\mathcal{D}|})$ is a document, we shuffle the sentences in $\mathcal{D}$ and obtain a permuted document $\mathcal{D}' = (S'_{1}, S'_2, \dots, S'_{|\mathcal{D}|}) $ where $ S_{i} $ is the $i$th sentence in the original document and there exists a sentence $S'_{P_i} = S_{i}$ in the permuted document $\mathcal{D}'$ (i.e., $P_i \in [1, |\mathcal{D}|]$ is the position of $S_{i}$ in $\mathcal{D}'$). In this task, we predict $\mathcal{P}=(P_1, P_2, \dots, P_{|\mathcal{D}|})$. 

As shown in Figure \ref{fig:reordering_architecture}, we first use the document encoder in Section \ref{sec:docmodel} to encode $\mathcal{D}'$ and yields its context dependent sentence representations $\mathbf{H}'=(\mathbf{h}'_1, \mathbf{h}'_2, \dots, \mathbf{h}'_{|\mathcal{D}|})$. Supposing that $P_0, P_1, P_2, \dots, P_{t-1}$ are known\footnote{We set $P_0 = 0$; $\mathbf{h}'_0$ is a zero vector.}, we  predict $P_t$ using a Pointer Network \cite{vinyals:2015:nips} with Transformer as its decoder. Let $TransDec^P$  denote the transformer decoder in \mbox{PointerNet},  $\mathbf{E}_{P_{i}}$ is the \emph{absolute} positional embedding of $P_i$ in original document and $\mathbf{p}_i$ the positional embedding of $P_i$ during decoding. The input of $TransDec^P$ is the sum of sentence representations and positional embeddings:
\begin{equation*}
\mathbf{M}_{t-1} \!=\! (\mathbf{h}'_{P_1} + \mathbf{p}_1 +\mathbf{E}_{P_1}, \dots, \mathbf{h}'_{P_{t-1}} + \mathbf{p}_{t-1} + \mathbf{E}_{P_{t-1}})
\end{equation*}
The output $\mathbf{h}^o_t$ summaries the sentences \emph{de-permutated} so far.
\begin{equation}
\mathbf{h}^o_t = TransDec^P( \mathbf{M}_{t-1} )
\end{equation}
Then the probability of selecting $S'_{p_t}$ is estimated with the attention \cite{bahdanau:2015:iclr} between $\mathbf{h}^o_t$ and all sentences in $\mathcal{D}'$ as follows:
\begin{equation}
p(P_t | P_{1:t-1}, \mathcal{D}') = \frac{\exp(g(\mathbf{h}^o_t, \mathbf{h}'_{P_t}))}{\sum_i \exp( g(\mathbf{h}^o_t, \mathbf{h}'_{i}) )}
\end{equation}
where $g$ is a feed forward neural network with the following parametrization:
\begin{equation}
g(\mathbf{h}^o_t, \mathbf{h}'_{i}) = \mathbf{v}_a^{\top} \tanh ( \mathbf{U}_a  \mathbf{h}^o_t +  \mathbf{W}_a  \mathbf{h}'_{i} )
\end{equation}
where $\mathbf{v}_a \in \mathbb{R}^{d \times 1}$, $\mathbf{U}_a \in \mathbb{R}^{d \times d}$, $\mathbf{W}_a \in \mathbb{R}^{d \times d}$ are trainable parameters.
Finally the probability of positions of original sentences in the shuffled document is:
\begin{equation}
\label{eq:pro_shuffling}
p(\mathcal{P} | \mathcal{D}') = \prod_{t=1}^{|\mathcal{D}|} p(P_t | P_{0:t-1}, \mathcal{D}')
\end{equation}

During training, for each batch of documents we apply both the masked sentence prediction and sentence shuffling tasks. One document $\mathcal{D}$ generates a masked document $\widetilde{\mathcal{D}}$ and a shuffled document $\mathcal{D}'$. Note that 15\% of sentences are masked in the masked document $\widetilde{\mathcal{D}}$, and all sentences are shuffled in the shuffled document $\mathcal{D}'$. The whole model is optimized with the following objective:
\begin{equation*}
L(\theta) = - \sum_{\mathcal{D} \in \mathcal{X}} \log p( \mathcal{O} |  \widetilde{\mathcal{D}} ) + \log p(\mathcal{P} | \mathcal{D}')
\end{equation*}
where $\mathcal{D}$ is a document in the training document set $\mathcal{X}$.

\subsection{Unsupervised Summarization}
\label{sec:extract_sum}

In this section, we propose our unsupervised extractive summarization method. Extractive summarization aims to select the most important sentences in document. 
Once we have obtained a hierarchical encoder using the pre-training methods in Section \ref{sec:pretraining}, we are ready to rank sentences and no additional fine-tuning is needed in this step. 

Our first ranking criteria is based on the probabilities of sentences in a document. Recall that $\mathcal{D} = ( S_1, S_2, \dots, S_{| \mathcal{D} |} )$ is a document and its probability is 
\begin{equation}
p( \mathcal{D} ) = \prod_{i=1}^{ | \mathcal{D} | } p(S_i | S_{1:i-1}) \approx \prod_{i=1}^{ | \mathcal{D} | } p(S_i | \mathcal{D}_{\neg S_i})
\end{equation}
It is not straight forward to estimate $p(S_i | S_{1:i-1})$ directly since document models in this work are all bidirectional. However, we can estimate  $p(S_i | \mathcal{D}_{\neg S_i})$ using the masked sentences prediction task in Section \ref{sec:pretraining}. We therefore use $p(S_i | \mathcal{D}_{\neg S_i})$ to approximate $p(S_i | S_{1:i-1})$. Finding the most important sentence is equivalent to finding the sentence with highest probability (i.e., $p(S_i | \mathcal{D}_{\neg S_i})$). In the following we demonstrate how to estimate $p(S_i | \mathcal{D}_{\neg S_i})$. As in Section \ref{sec:pretraining}, we create $\mathcal{D}_{\neg S_i}$
 by masking $S_i$ in $\mathcal{D}$ (i.e., replacing all tokens in $S_i$ with {\tt [MASK]} tokens). $p(S_i | \mathcal{D}_{\neg S_i})$ can be estimated using Equation (\ref{eq:pro_masked}). 
To make the probabilities of different sentences comparable, we normalize them by their length. Then we obtain $\hat{ r }_i$ as follows\footnote{We also tried the geometric average, but the effect is not as good as the arithmetic average.} (also see Equation (\ref{eq:pro_masked}))
\begin{equation}
\hat{ r }_i = \frac{1}{ |S_i| } \sum_{j=1}^{\rvert S_i \rvert} p(w_j^i | w_{0:j-1}^i, \mathcal{D}_{\neg S_i})
\end{equation}
We also normalize $\hat{r}_i$ across sentences (in a document) and obtain our first ranking criteria $\widetilde{r}_i$:
\begin{equation}
    \widetilde{r}_i = \frac{\hat{r}_i}{\sum_{j=1}^{|\mathcal{D}|}\hat{r}_j} 
\end{equation}

In the second ranking criteria, we model the contributions of other sentences to the current sentence explicitly. We view a document $\mathcal{D}$ as a directed graph, where each sentence in it is a node. The connections between sentences (i.e., edge weights) can be modeled using the self-attention matrix $\mathbf{A}$ of the sentence level Transformer encoder described in Section \ref{sec:docmodel}, which is produced by a pre-trained hierarchical document encoder. We assume that a sentence $S_j$ can transmit its importance score $\widetilde{ r }_i $ to an arbitrary sentence $S_i$ through the edge between them. Let $\mathbf{A}_{j,i}$ denote the attention score from $S_j$ to $S_i$. After receiving all transmissions from all sentences, the second ranking score for $S_i$ is as follows:

\begin{equation}
\label{eq:attn_score}
r'_i = \sum_{j=1, j\neq i}^{| \mathcal{D} |} \mathbf{A}_{j,i}  \times \widetilde{ r }_j
\end{equation}


The final ranking score of $S_i$ combines the score from itself as well as other sentences:
\begin{equation}
\label{eq:rank}
r_i = \gamma_1 \: \widetilde{ r }_i  + \gamma_2 \: r'_i
\end{equation}
$\gamma_1$ and $\gamma_2$ are coefficients tuned on development set. $r_i$ can be computed iteratively by assigning $r_i $ to $\widetilde{r}_i$ and repeating Equation (\ref{eq:attn_score}) and Equation (\ref{eq:rank}) for $T$ iterations. We find a small $T$ ($T \le 3$) works well according to the development set.

\section{Experiments}
In this section we assess the performance of \textsc{Stas} on the document summarization task. We firstly introduce datasets we used and then give our implementation details. Finally we compare our method against previous methods. 
\subsection{Datasets}
We evaluate \textsc{Stas} on two summarization datasets, namely the CNN/DailyMail (CNN/DM; \citealt{hermann:2015:nips}) dataset and the New York Times (NYT; \citealt{sandhaus:2008:ldc}) dataset. CNN/DM is composed of articles from CNN and Daily Mail news websites, which uses their associated highlights as reference summaries. NYT dataset contains articles published by the New York Times between January 1, 1987 and June 19, 2007 and summaries are written by library scientists.
For the CNN/DM dataset, we follow the standard splits and pre-processing steps used in supervised summarization\footnote{scripts available at https://github.com/nlpyang/PreSumm} \cite{see-etal-2017-get,liu-lapata-2019-text}, and the resulting dataset contains 287,226 articles for training, 13,368 for validation and 11,490 for test. 
Following \newcite{zheng-lapata-2019-sentence}, we adopted the splits widely used in abstractive summarization \cite{paulus2018a} for the NYT dataset, which ranks articles by their publication date and used the first 589,284 for training, the next 32,736 for validation and the remaining 32,739 for test. Then, we filter out documents whose summaries are shorter than 50 words as in \cite{zheng-lapata-2019-sentence} and finally retain 36,745 for training, 5,531 for validation and 4,375 for test. 

We segment sentences using the Stanford CoreNLP toolkit \cite{manning-EtAl:2014:P14-5}. Sentences are then tokenized with the UTF-8 based BPE tokenizer used in RoBERTa and GPT-2 \cite{radford2019language} and the resulting vocabulary contains 50,265 subwords. During training, we only leverage articles in CNN/DM or NYT; while we do use both articles and summaries in validation sets to tune hyper-parameters of our models. 

We evaluated the quality of summaries from different models using ROUGE \cite{lin-2004-rouge}. We report the full length F1 based ROUGE-1, ROUGE-2, ROUGE-L on both CNN/DM and NYT datasets. These ROUGE scores are computed using the {\tt ROUGE-1.5.5.pl} script\footnote{https://shorturl.at/nAG49}. 

\subsection{Implementation Details}

The main building blocks of \textsc{Stas} are Transformers \cite{vaswani:2017:nips}. In the following, we describe the sizes of them using the number of layers $L$, the number of attention heads $A$, and the hidden size $N$. As in \cite{vaswani:2017:nips,devlin:2019:naacl}, the hidden size of the feed-forward sublayer is always $4H$. \textsc{Stas} contains one hierarchical encoder (see Section \ref{sec:docmodel}) and two decoders, where they are used for the masked sentences prediction and sentence shuffling pre-training tasks (see Section \ref{sec:pretraining}). The token-level encoder is initialized with the parameters of $\text{RoBERTa}_{\text{BASE}}$ \cite{liu:2019:roberta}\footnote{We also tried  $\text{RoBERTa}_{\text{LARGE}}$ and obtained worse results.} and we set $L=12, H=768, A=12$. The sentence-level encoder and the two decoders are shallower and we all adopt the setting $L=6, H=768, A=12$. 

%

We trained our models with 4 Nvidia Tesla V100 GPUs and optimized them using Adam \cite{Kingma2015AdamAM} with $\beta_1 = 0.9, \beta_2 = 0.999$. Since the encoder is partly pre-trained (initialized with RoBERTa) and the decoders are initialized randomly, we set a lager learning rate for decoders. Specifically, we used 4e-5 for the encoder and 4e-4 for the decoders. Since CNN/DM is larger than NYT, we employed a batch size of 512 for CNN/DM and 64 for NYT (to ensure a sufficient number of model updates)\footnote{We used gradient accumulation technique \cite{ott-etal-2019-fairseq} to increase the actual batch size.}. Limited by the positional embedding of RoBERTa, all documents are truncated to 512 subword tokens. We trained our models on both CNN/DM and NYT for 100 epochs. It takes around one hour training on the CNN/DM and 30 minutes on the NYT for each epoch. The best checkpoint is at around epoch 85 on CNN/DM  and epoch 65 on NYT according to validation sets.



When extracting the summary for a new document during test time, we rank all sentences using Equation (\ref{eq:rank}) and select the top-3 sentences as the summary. 
When selecting sentences on the CNN/DM dataset, we find that \textit{trigram blocking} (i.e., removing sentences with repeating trigrams to existing summary sentences) \cite{paulus2018a} can reduce the redundancy, while \textit{trigram blocking} does not help on NYT.

%


\begin{table*}
	\small
	\begin{center}
		\begin{tabular}{|l | r r r  | r r r |}
			\hline
			\multirow{2}{*}{\textbf{Method}}  & \multicolumn{3}{c|}{\textbf{CNN/DM}} & \multicolumn{3}{c|}{\textbf{NYT}}\\
			&R-1 & R-2 & R-L & R-1 & R-2 &R-L \\
			\hline
			\hline
			REFRESH \cite{narayan-etal-2018-ranking} & 41.30 & 18.40 & 37.50 & 41.30 & 22.00 & 37.80 \\
			PTR-GEN \cite{see-etal-2017-get} & 39.50 & 17.30 & 36.40& \textbf{42.70} & \textbf{22.10}& \textbf{38.00} \\
			BertSumExt \cite{liu-lapata-2019-text} & \textbf{43.25} & \textbf{20.24} & \textbf{39.63} & \multicolumn{1}{c}{--} & \multicolumn{1}{c}{--} & \multicolumn{1}{c|}{--}  \\
			BertSumAbs \cite{liu-lapata-2019-text} & 41.72 & 19.39 & 38.76 & \multicolumn{1}{c}{--} & \multicolumn{1}{c}{--} & \multicolumn{1}{c|}{--}  \\
			\hline
			\hline
			LEAD-3 & 40.50 & 17.70 & 36.70 & 35.50 & 17.20 & 32.00 \\
			TEXTRANK (tf-idf) & 33.20 & 11.80 & 29.60 & 33.20 & 13.10 & 29.00 \\
			TEXTRANK (skip-thought) & 31.40& 10.20& 28.20& 30.10& 9.60& 26.10 \\
			TEXTRANK (BERT) & 30.80 & 9.60 & 27.40 & 29.70 & 9.00 & 25.30 \\
			PACSUM \cite{zheng-lapata-2019-sentence} & 40.70 & 17.80 & 36.90 & 41.40 & 21.70 & 37.50 \\
			PACSUM (BERT) * &40.69 &17.82 &36.91 & 40.67 & 21.09 &36.76 \\
			PACSUM (RoBERTa) * &40.74 &17.82 &36.96 &40.84 &21.28 &37.03 \\
			Adv-RF \cite{wang-lee-2018-learning} &35.51& 9.38& 20.98& \multicolumn{1}{c}{--} & \multicolumn{1}{c}{--} & \multicolumn{1}{c|}{--} \\
			TED \cite{Yang2020TEDAP}  & 38.73 & 16.84 & 35.40 & 37.78 & 17.63 & 34.33 \\
			\hline \hline
			\textsc{Stas}  & \textbf{40.90} &\textbf{18.02} & \textbf{37.21}& \textbf{41.46} & \textbf{21.80} & \textbf{37.57} \\
			
			\textsc{Stas} + PACSUM & \textbf{41.26} &\textbf{18.18} &\textbf{37.48} & \textbf{42.42}&\textbf{22.66} & \textbf{38.50}\\
			\hline
		\end{tabular}
	\end{center}
	\caption{Results on CNN/DM and NYT test sets using ROUGE F1. * means our own re-implementation. Results of TEXTRANK (tf-idf), TEXTRANK (skip-thought) and TEXTRANK (BERT) are reported in \cite{zheng-lapata-2019-sentence} and they use tf-idf, skip-thought vectors \cite{kiros2015skip} and BERT as sentence features, respectively.
	}
	\label{tab:main_result}
\end{table*}

\subsection{Results}
Our main results are shown in Table \ref{tab:main_result}.
The first block includes several recent supervised models for document summarization. REFRESH \cite{narayan-etal-2018-ranking} is an extractive model, which is trained by globally optimizing the ROUGE metric with reinforcement learning. PTR-GEN \cite{see-etal-2017-get} is a sequence to sequence based abstractive model with copy and coverage mechanism. \newcite{liu-lapata-2019-text} initialize encoders of extractive model (BertSumExt) and abstractive model (BertSumAbs) with pre-trained BERT.  

We present the results of previous unsupervised methods in the second block. LEAD-3 simply selects the first three sentences as the summary for each document. TEXTRANK \cite{mihalcea2004textrank} views a document as a graph with sentences as nodes and edge weights using the sentence similarities. It selects top sentences as summary w.r.t. PageRank \cite{page:1999:pagerank} scores. PACSUM \cite{zheng-lapata-2019-sentence} is yet another graph-based extractive model using BERT as sentence features. Sentences are ranked using centralities (sum of all out edge weights). They made the ranking criterion positional sensitive by forcing negative edge weights for edges between the current sentence and its preceding sentences.
\mbox{Adv-RF} \cite{wang-lee-2018-learning} and TED \cite{Yang2020TEDAP} are all based on unsupervised seq2seq auto-encoding with additional objectives of adversarial training, reinforcement learning and seq2seq pre-training to predict leading sentences. 

PACSUM is based on the BERT \cite{devlin:2019:naacl} initialization. RoBERTa \cite{liu:2019:roberta}, which extends BERT with better training strategies and more training data, outperforms BERT on many tasks. We therefore re-implemented PACSUM and extended it with both BERT and RoBERTa initialization (i.e., PACSUM (BERT) and PACSUM (RoBERTa))\footnote{We re-implemented PACSUM, because the training code of PACSUM is not available.}. On CNN/DM, our re-implementation \mbox{PACSUM (BERT)} is comparable with \newcite{zheng-lapata-2019-sentence}. The results of \mbox{PACSUM (BERT)} and the RoBERTa initialized \mbox{PACSUM (RoBERTa)} are almost the same. Perhaps because it relies more on position information rather than sentence similarities computed by BERT or RoBERTa. 
\textsc{Stas} outperforms all unsupervised models in comparison on CNN/DM and the difference between \textsc{Stas} and all other unsupervised models are significant with a 0.95 confidence interval according to the ROUGE script. In the following, all significant tests on ROUGE are measured with a 0.95 confidence interval using the ROUGE script. Since \textsc{Stas} does not model sentence positions explicitly during ranking, while PACSUM does, we linearly combine the  ranking scores of \textsc{Stas} and PACSUM (i.e., \textsc{Stas} + PACSUM)\footnote{We first normalize sentence scores in each document for both \textsc{Stas} and PACSUM. In the combination, weight for \textsc{Stas} is 0.9 and weight for PACSUM is 0.1 (tuned on validation sets).}. The combination further improves the performance. 

On NYT, the trend is similar. \textsc{Stas} is slightly better than PACSUM although not significantly better (\textsc{Stas} is siginificantly better than all the other unsupervised models in comparison). Interestingly, there are also no significant differences between \textsc{Stas} and two supervised models (REFRESH and PTR-GEN). \mbox{\textsc{Stas} + PACSUM} even significantly outperforms the supervised REFRESH. The significant tests above all utilize the ROUGE script. 

Examples of gold summaries and system outputs of REFRESH \cite{narayan-etal-2018-ranking}, STAS and PACSUM \cite{zheng-lapata-2019-sentence} on the CNN/DM dataset can be found in  appendix B.


\begin{table}
	\scriptsize
	\begin{center}
		\begin{tabular}{|l | c c c  | c c c|}
			\hline
			\multirow{2}{*}{\textbf{Settings}}  & \multicolumn{3}{c|}{\textbf{valid set}} & \multicolumn{3}{c|}{\textbf{test set}}\\
			&R-1 & R-2 & R-L & R-1 & R-2 &R-L \\
			\hline
			MSP & 41.61 &18.30 & 37.92 & 40.76 & 17.78 & 37.03 \\
			MSP+SS (\textsc{Stas}) &41.67 &18.47& 38.00& 40.90 & 18.02 & 37.21 \\
			\hline
			$\widetilde{ r } = 1/ |\mathcal{D}|$ &41.58& 18.43&37.89 &40.74 &17.88 &37.04 \\
			$r'=0$ &33.92& 12.93& 30.99 &33.30 &12.61 &30.33  \\
			\hline
		\end{tabular}
		\caption{Ablation study on CNN/DM validation and test sets using ROUGE F1. 
		}
		\label{tab:ablation_result}
	\end{center}
\end{table}

\subsection{Analysis}

\paragraph{Ablation Study}
In Section \ref{sec:pretraining}, we proposed two pre-training tasks. \emph{Are they are all useful for summarization?} As shown in Table \ref{tab:ablation_result}, when we only employ the masked sentences prediction task (MSP), we can obtain a ROUGE-2 of 17.73, which is already very close to the result of PACSUM (see Table \ref{tab:main_result}). When we add the sentence shuffling task (denoted as MSP+SS (\textsc{Stas})), we improves the performance over MSP. Note that we can not use only the sentence shuffling task (SS), because the first term in our sentence scoring equation (see Equation (\ref{eq:rank})) depends on the probabilities produced by decoder in the MSP task. 


%

In section \ref{sec:extract_sum}, we propose two criteria to score sentences (see the two terms in Equation (\ref{eq:rank})). The effects of them are shown in the second block of Table \ref{tab:ablation_result}.
 Since the attention based criterion $r'$ relies on sentence probability based criterion $\widetilde{r}$, we cannot remove  $\widetilde{r}$ and instead we set $\widetilde{r} = \frac{1}{\mathcal{D}}$ to see the effect of $\widetilde{r}$. As a result, ROUGE-2 decreases by 0.14, which indicates that $\widetilde{r}$ is necessary for ranking.
 Also note that when setting $\widetilde{r} = \frac{1}{\mathcal{D}}$, our method is equivalent to PageRank using sentence level attention scores as edge weights. Instead of iterating until convergence as in the original PageRank algorithm, we find a small iteration number ($T \le 3$) is sufficient.  To study the effect of the attention based criterion $r'$, we set $r'=0$, which means sentences are ranked using sentence probability based criterion $\widetilde{r}$. We can see that the performance drops dramatically by 5 ROUGE-2. 
\begin{table}
	\scriptsize
	\begin{center}
		\begin{tabular}{|l | c c c  | c c c|}
			\hline
			\multirow{2}{*}{\textbf{}}  & \multicolumn{3}{c|}{\textbf{valid set}} & \multicolumn{3}{c|}{\textbf{test set}}\\
			&R-1 & R-2 & R-L & R-1 & R-2 &R-L \\
			\hline
			 w/ $A_{i,j}$  & 33.66 & 12.78 & 30.75 & 33.02 & 12.48 & 30.08  \\
			w/ $A_{j, i}$  & 41.67 & 18.47 & 38.00 & 40.90 & 18.02 & 37.21 \\
			\hline
		\end{tabular}
		\caption{$\mathbf{A_{j,i}}$ v.s. $\mathbf{A_{i,j}}$ on CNN/DM validation and test sets using ROUGE F1.}
		\label{tab:Transpose}
	\end{center}
\end{table}
\paragraph{$\mathbf{A_{j,i}}$ v.s. $\mathbf{A_{i,j}}$}
In Equation (\ref{eq:attn_score}), we compute $r_i'$ with $\mathbf{A}_{j,i}$ (attention score from $S_j$ to $S_i$). The intuition of using $\mathbf{A}_{j,i}$ is that a sentence is important if the interpretation of other important sentences depends on it. However, an alternative is to use $\mathbf{A}_{i, j}$. It shows in Table \ref{tab:Transpose} that $\mathbf{A}_{j,i}$ is indeed better.

\begin{figure}[t]
	\centering
	\includegraphics[width=1\linewidth]{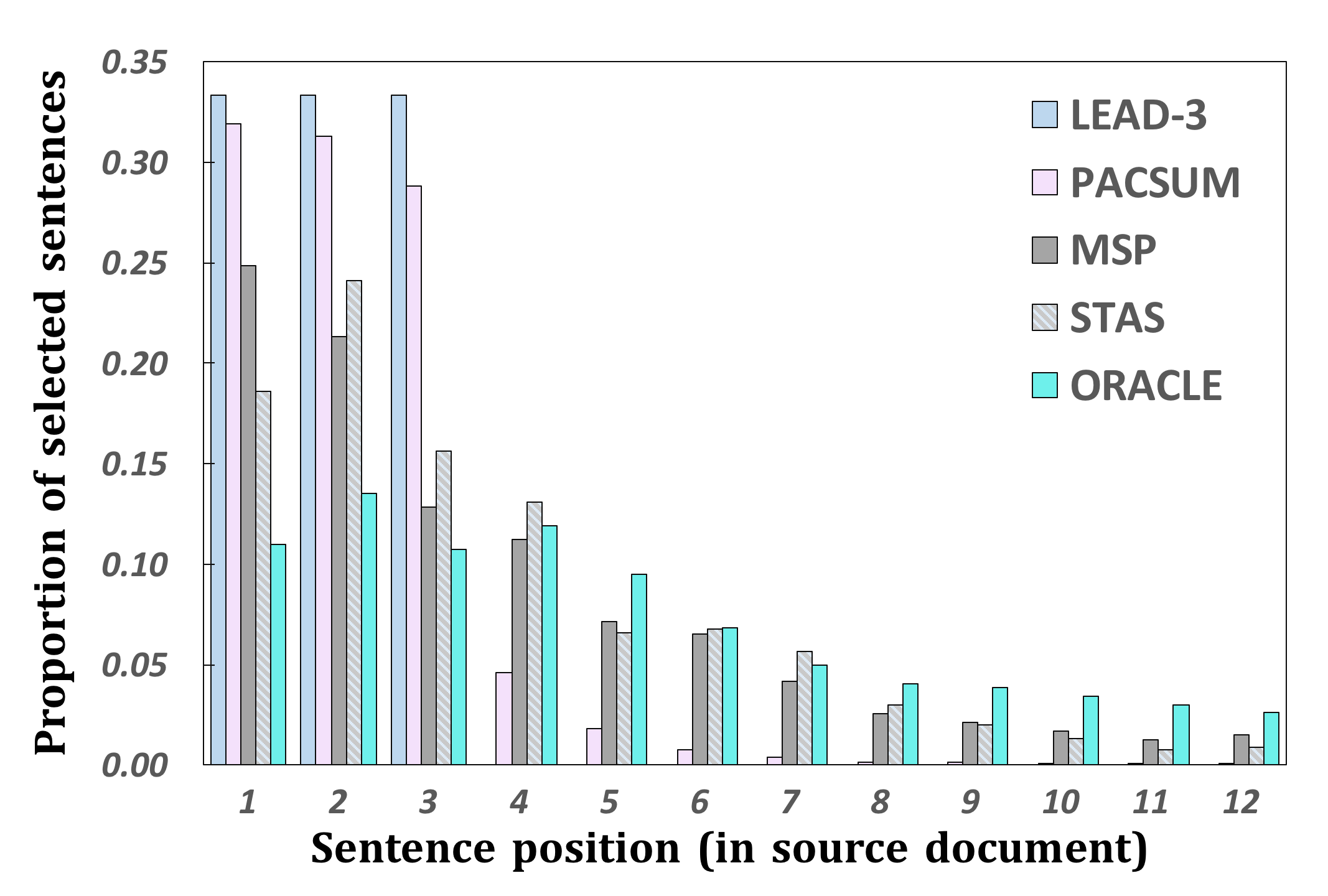}
	\caption{Proportion of extracted sentences by different unsupervised models against their positions. }  
	\label{fig:pos_compare} 
\end{figure}

\paragraph{Sentence Position Distribution}
We also analyze \emph{how extractive sentences by different models are distributed in documents?} We compare \textsc{Stas} against LEAD-3, PACSUM and ORACLE using the first 12 sentences in all documents on CNN/DM test set. ORACLE is the upper bound for extractive models. Extractive summaries of ORACLE are generated by selecting a subset of sentences in a document, which maximize ROUGE score \cite{nallapati2017summarunner}. 
As shown in Figure 4, we can see that sentences selected by ORACLE are smoothly distributed across all positions,  while LEAD-3 only selects the first 3 sentences. Compared to \textsc{Stas}, the sentence distribution of PACSUM is closer to that of LEAD-3 and \textsc{Stas} produces a sentence distribution that is more similar to that of ORACLE. The observation above indicates that our model relies less on sentence positions compared to PACSUM. We further computed the Kullback Leibler divergence between the sentence position distribution of an unsupervised model and the  distribution of ORACLE and we denote it as $KL(\cdot || \text{ORC})$. We found $KL(\text{PACSUM} || \text{ORC}) =0.614 $ is much large than $KL(\textsc{Stas} || \text{ORC}) = 0.098$, indicating \textsc{Stas} is better correlated with ORACLE. We introduce the sentence shuffling task to encourage \textsc{Stas} to select sentences based on their contents rather than their positions only (see Section \ref{sec:pretraining}).  After we remove the sentence shuffling task from \textsc{Stas} during pre-training (see MSP in Figure \ref{fig:pos_compare}), there is a clear trend that leading sentences are more frequently selected. Moreover,  $KL(\textsc{Stas} || \text{ORC}) <  KL(\text{MSP} || \text{ORC}) = 0.108$. By introducing the sentence shuffle task, sentence positional distribution of \textsc{Stas} is closer to that of ORACLE.  
\begin{table}[ht]
	\scriptsize
	\begin{center}
		\begin{tabular}{|l | c c c  | c c c|}
			\hline
			\multirow{2}{*}{\textbf{}}  & \multicolumn{3}{c|}{\textbf{valid set}} & \multicolumn{3}{c|}{\textbf{test set}}\\
			&R-1 & R-2 & R-L & R-1 & R-2 &R-L \\
			\hline
			MSP& 41.61 & 18.30 & 37.92 & 40.76 & 17.78 & 37.03 \\
			MSP+(a)  & 40.93 & 17.73 & 37.27 & 40.15 & 17.25 & 36.46  \\
 			MSP+(b) & 37.59 & 14.71 & 33.95 & 36.91& 14.26 & 33.25 \\
			MSP+SS  & 41.67 & 18.47 & 38.00 & 40.90 & 18.02 & 37.21 \\
			\hline
		\end{tabular}
		\caption{Compare SS with other two different methods which remove sentence positions. (a) remove the sentence level positional embedding; (b) for each sentence in the token level, use a positional embedding from positional 0. Results are reported on  CNN/DM.}
		\label{tab:ss_compare}
	\end{center}
\end{table}
\paragraph{Why Sentence Shuffling?} Since the Sentences Shuffling task aims to make {\sc Stas} less dependent on sentence positions. However, there are potentially simpler methods to remove sentence position information. For example, (a) we can remove the sentence-level positional embedding and (b) for each sentence in the token level, we can use a positional embedding from positional 0. Results in Table \ref{tab:ss_compare} indicates that upon the MSP objective, strategies (a) and (b) hurt the performance of MSP significant, while SS improves over MSP. It may because positional embeddings, whether on token or sentence level, are important (at least for the MSP task). One advantage of SS over (a) and (b) is that it can make our model less dependent on positions (see Section 4.4) and retain the power of positional embeddings and the MSP objective at the same time.

\section{Conclusions}
In this paper, we find that (sentence-level) transformer attention (in a hierarchical transformer) can be used to rank sentences for unsupervised extractive summarization, while previous work leverage graph based (or rule based) methods and sentence similarities computed with off-the-shelf sentence embeddings. We propose the sentence shuffling task for pre-training hierarchical transformers, which helps our model to select sentences based on their contents rather than their positions only. 
Experimental results on CNN/DM and NYT datasets show that our model outperforms other recently proposed unsupervised methods. The sentence position distribution analysis shows that our method is less dependent on sentence positions. When combined with recent unsupervised model explicitly modeling sentence positions, we obtain even better results. In the next step, we plan to apply our models to unsupervised abstractive summarization. 

\section*{Acknowledgments}

We gratefully acknowledge Hao Zheng for the technical advice during our  re-implementation of \mbox{PACSUM} \cite{zheng-lapata-2019-sentence}. We would also like to thank the anonymous reviewers for their insightful feedback.

\bibliography{emnlp2020}
\bibliographystyle{acl_natbib}

\newpage
\appendix


\section{Results on the validation set }
We also provide the results on the validation set when combining \textsc{Stas} with PACSUM in Table \ref{tab:ensemble}. The result shows that \textsc{Stas} and PACSUM evaluate the sentences from different perspectives.

\begin{table}[ht]
	\scriptsize
	\setlength{\belowcaptionskip}{-0.8cm}
	\begin{center}
		\begin{tabular}{|l | c c c  | c c c|}
			\hline
			\multirow{2}{*}{\textbf{Methods}}  & \multicolumn{3}{c|}{\textbf{valid set}} & \multicolumn{3}{c|}{\textbf{test set}}\\
			&R-1 & R-2 & R-L & R-1 & R-2 &R-L \\
			\hline
			\multicolumn{7}{|c|}{\textbf{CNN/DM}} \\
			\hline
			PACSUM & - & -& -& 40.70 & 17.80 & 36.90 \\
			\textsc{Stas} & 41.67 &18.47& 38.00 & 40.90 & 18.02 & 37.21 \\
			\textsc{Stas} + PACUSM &42.20 &18.84& 38.44& 41.26 & 18.18 & 37.48 \\
			\hline
			\multicolumn{7}{|c|}{\textbf{NYT}} \\
			\hline
			PACSUM & - & - & - & 41.40 & 21.70 & 37.50 \\
		    \textsc{Stas} &40.36& 20.20&36.00 &41.46 &21.80 &37.57 \\
			\textsc{Stas} + PACUSM &41.46& 21.22& 37.05 &42.42 &22.66 &38.50  \\
			\hline
		\end{tabular}
		\caption{Results of the combination using ROUGE F1 for CNN/DM and NYT}
		\label{tab:ensemble}
	\end{center}
\end{table}

\section{Examples}\label{example}
We show some example gold summaries and system outputs of REFRESH\cite{narayan-etal-2018-ranking}, \textsc{Stas} and PACSUM \cite{zheng-lapata-2019-sentence} on the CNN/DM dataset in table \ref{tab:example_1to2} and \ref{tab:example_3to4}.
\begin{table*}[ht]
    \centering
    \begin{tabular}{m{0.5cm}<{\centering}  m{6.5cm}  m{6.5cm}}
         & Article 1 & Article 2\\
        \hline
        \rotatebox{90}{GOLD} & Filmmaker Michael König from Cologne, Germany has created an amazing video showing solar activity. It was made by stitching together footage from Nasa's Solar Dynamics Observatory over five years. The footage includes loops of `coronal rain' showering the surface of the sun. Transits of the moon, Venus and Earth are also seen - and a comet breezes through the outer solar atmosphere.
        &Patrick Kramer, 33, is a hyper-realist artist from Springville, Utah. Works from photos and builds up layers of oil paints on canvas. Each work of art takes anywhere between 50 to 300 hours to complete.\\
        \hline
        \rotatebox{90}{REFRESH} & An incredible video has stitched together footage from a Nasa spacecraft, revealed the beauty of the solar surface as it bursts with energy. Taken over five years, the footage includes plasma raining down on the sun, an extreme solar eruption and even a comet breezing through the sun's atmosphere. The movie, called Sun, was created by filmmaker Michael König from Cologne, Germany. It uses footage recorded by Nasa's Solar Dynamics Observatory ( SDO ) spacecraft between 2011 and 2015. 
        &They become even more awe-inspiring at the realisation that they're paintings, and someone spent between 50 to 300 hours painstakingly creating them. Patrick Kramer is a 33-year-old artist hailing from Springville, Utah, who specialises in remarkably hyper-realistic paintings. One of his images is an imitation of the cover art for musician Drake's 2013 album, Nothing Was the Same. The US artist works from photos and builds up delicate layers of oil paint on a board or canvas.
        \\
        \hline
        \rotatebox{90}{\textsc{Stas}} &An incredible video has stitched together footage from a Nasa spacecraft, revealed the beauty of the solar surface as it bursts with energy. Filmmaker Michael König from Cologne, Germany has created an amazing video showing solar activity ( shown ). It uses footage recorded by Nasa's Solar Dynamics Observatory ( SDO ) spacecraft between 2011 and 2015. 
        &Patrick Kramer is a 33-year-old artist hailing from Springville, Utah, who specialises in remarkably hyper-realistic paintings. The US artist specialises in beautifully in-depth oil painted portraits, including this intricate black and white portrait of a man called Matt. It's almost impossible to tell that Kramer's flawless creations are in fact oil paintings due to their incredible pin-sharp detail and clarity.\\
        \hline 
        \rotatebox{90}{PACSUM} & You might have seen fantastic images of the sun before, or even clips showing its activity - but you've never seen anything like this. An incredible video has stitched together footage from a Nasa spacecraft, revealed the beauty of the solar surface as it bursts with energy. Taken over five years, the footage includes plasma raining down on the sun, an extreme solar eruption and even a comet breezing through the sun's atmosphere.
        &Love the lighting in these photography shots? They become even more awe-inspiring at the realisation that they're paintings, and someone spent between 50 to 300 hours painstakingly creating them. Patrick Kramer is a 33-year-old artist hailing from Springville, Utah, who specialises in remarkably hyper-realistic paintings.\\
    \end{tabular}
    \caption{Example gold summaries and system outputs of REFRESH\cite{narayan-etal-2018-ranking}, \textsc{Stas} and PACSUM \cite{zheng-lapata-2019-sentence} on the CNN/DM test dataset}
    \label{tab:example_1to2}
\end{table*}

\begin{table*}[ht]
    \centering
    \begin{tabular}{m{0.5cm}<{\centering}  m{6.5cm}  m{6.5cm}}
         & Article 3 & Article 4\\
        \hline
        \rotatebox{90}{GOLD}
        &AQAP says a ``crusader airstrike'' killed Ibrahim al-Rubaish. Al-Rubaish was once detained by the United States in Guantanamo.
        &Queen Victoria's holiday residence was Osborne House on the Isle of Wight. But her journeys there involved train and ferry ride and then another train ride to a station more than two miles from the property. In 1875, a station was built at Whippingham just to serve Royal residence. Building is now a five-bedroom home, currently on the market for 625,000.\\
        \hline
        \rotatebox{90}{REFRESH}
        &A top al Qaeda in the Arabian Peninsula leader -- who a few years ago was in a U.S. detention facility -- was among five killed in an airstrike in Yemen, the terror group said, showing the organization is vulnerable even as Yemen appears close to civil war. Ibrahim al-Rubaish died Monday night in what AQAP's media wing, Al-Malahem Media, called a ``crusader airstrike.'' Al-Rubaish was once held by the U.S. government at its detention facility in Guantanamo Bay, Cuba.
        &` It is impossible to imagine a prettier spot,' Queen Victoria said of her holiday residence, Osborne House on the Isle of Wight. It was at the beach there that she would take to the sea in her wooden bathing machine, and that all of her children learned to swim. She would travel to Portsmouth by train and then by ferry to Ryde. So, in 1875, a station was built at Whippingham, the closest point on the line to Osborne House -- just to serve the Royal residence.\\
        \hline
        \rotatebox{90}{\textsc{Stas}}
        &A top al Qaeda in the Arabian Peninsula leader -- who a few years ago was in a U.S. detention facility -- was among five killed in an airstrike in Yemen , the terror group said , showing the organization is vulnerable even as Yemen appears close to civil war. Ibrahim al-Rubaish died Monday night in what AQAP 's media wing , Al-Malahem Media , called a `` crusader airstrike . " Al-Rubaish was once held by the U.S. government at its detention facility in Guantanamo Bay , Cuba.
        &Pretty spot : Queen Victoria spent her holidays in Osborne House on the Isle of Wight. So , in 1875 , a station was built at Whippingham , the closest point on the line to Osborne House -- just to serve the Royal residence. The current owners , retired electrician Tony Vickers , 69 , and his 70-year-old wife Sue , a retired hairdresser , are selling because they want to move to a smaller bungalow .\\
        \hline 
        \rotatebox{90}{PACSUM}
        &A top al Qaeda in the Arabian Peninsula leader -- who a few years ago was in a U.S. detention facility -- was among five killed in an airstrike in Yemen, the terror group said, showing the organization is vulnerable even as Yemen appears close to civil war. Ibrahim al-Rubaish died Monday night in what AQAP's media wing, Al-Malahem Media, called a ``crusader airstrike.'' The Al-Malahem Media obituary characterized al-Rubaish as a religious scholar and combat commander.
        &` It is impossible to imagine a prettier spot,' Queen Victoria said of her holiday residence, Osborne House on the Isle of Wight. It was at the beach there that she would take to the sea in her wooden bathing machine, and that all of her children learned to swim. But, however pleasing her days at Osborne House, Victoria’s journeys there were never easy. \\
    \end{tabular}
    \caption{Example gold summaries and system outputs of REFRESH\cite{narayan-etal-2018-ranking}, \textsc{Stas} and PACSUM \cite{zheng-lapata-2019-sentence} on the CNN/DM test dataset}
    \label{tab:example_3to4}
\end{table*}

\end{document}